%% file: tacl.tex
\documentclass[11pt,letterpaper]{article}
\usepackage[letterpaper]{geometry}
\usepackage{acl2012}
\usepackage{natbib}
\usepackage{times}
\usepackage{latexsym}
\usepackage{amsmath}
\usepackage{multirow}
\usepackage{url}
\makeatletter
\newcommand{\@BIBLABEL}{\@emptybiblabel}
\newcommand{\@emptybiblabel}[1]{}
\makeatother
\usepackage[hidelinks]{hyperref}
\input{preamble}
\graphicspath{{fig/}}

\title{Learning with Latent Language}

\author{
  Jacob Andreas ~~ Dan Klein ~~ Sergey Levine \\
  Computer Science Division \\
  University of California, Berkeley \\
  {\tt \{jda,klein,svlevine\}@eecs.berkeley.edu }
}

\date{}

\begin{document}
\maketitle

\begin{abstract}
  \input{abstract}
\end{abstract}

\input{body}

\bibliography{jacob}
\bibliographystyle{acl_natbib}

\input{appendix}

\end{document}

%% file: preamble.tex
\usepackage{amsmath}
\usepackage{amsthm}
\usepackage{amssymb}
\usepackage{booktabs}
\usepackage{tikz}
\usepackage{multirow}
\usepackage[bb=fourier]{mathalfa}
\usepackage{inconsolata}
\usepackage{xspace}
\usepackage{wrapfig}
\usepackage{comment}
\usepackage{xcolor}
\usepackage{array}
\usepackage{arydshln}
\usepackage{makecell}

\DeclareMathOperator*{\argmin}{arg\,min}
\DeclareMathOperator*{\argmax}{arg\,max}
\newcommand{\reals}{\mathbb{R}}

\newcommand{\eg}{e.g.\ }
\newcommand{\ie}{i.e.\ }

\newcommand{\param}{\eta}
\newcommand{\paramspace}{\mathbf{H}}
\newcommand{\inp}{x}
\newcommand{\outp}{y}
\newcommand{\loss}{L}
\newcommand{\model}{f}

\newcommand{\tparam}{\theta}

\newcommand{\hint}{w}
\newcommand{\pinp}[2]{\inp^{(\ell #1)}_{#2}}
\newcommand{\poutp}[2]{\outp^{(\ell #1)}_{#2}}
\newcommand{\cinp}{\inp^{(c)}}
\newcommand{\coutp}{\outp^{(c)}}
\newcommand{\einp}{\inp^{(e)}}
\newcommand{\eoutp}{\outp^{(e)}}

\newcommand{\hparam}{\lambda}
\newcommand{\lc}{L$^3$\xspace}

\newcommand{\crep}{{\small\texttt{rep}}}
\newcommand{\cencode}{{\small\texttt{rnn-encode}}}
\newcommand{\cdecode}{{\small\texttt{rnn-decode}}}

%% file: abstract.tex
The named concepts and compositional operators present in natural language
provide a rich source of information about the kinds of abstractions humans use
to navigate the world. Can this linguistic background knowledge improve the
generality and efficiency of learned classifiers and control policies?  This
paper aims to show that using the space of natural language strings as a
\emph{parameter} space is an effective way to capture natural task structure.
In a pretraining phase, we learn a language interpretation model that transforms
inputs (\eg images) into outputs (\eg labels) given natural language
descriptions.  To learn a new concept (\eg a classifier), we search directly in
the space of descriptions to minimize the interpreter's loss on training
examples.  Crucially, our models do not require language data to learn these
concepts: language is used only in pretraining to impose structure on subsequent
learning.  Results on image classification, text editing, and reinforcement
learning show that, in all settings, models with a linguistic parameterization
outperform those without.\footnote{Code and data are available at
\url{http://github.com/jacobandreas/l3}.}

%% file: body.tex
\section{Introduction}
\label{sec:intro}

The structure of natural language reflects the structure of the world. For
example, the fact that it is easy for us to communicate the concept \emph{left
of the circle} but comparatively difficult to communicate \emph{mean saturation
of the first five pixels in the third column} reveals something about the
kinds of abstractions we find useful for interpreting and navigating our
environment \citep{Gopnik87WordLearning}.  In machine learning, efficient
automatic discovery of reusable abstract structure remains a major challenge.
This paper investigates whether background knowledge from language can provide
a useful scaffold for acquiring it.  We specifically propose to use language
as a latent \emph{parameter space} for few-shot learning problems of all
kinds, including classification, transduction and policy search.  We aim to
show that this linguistic parameterization produces models that are both more
accurate and more interpretable than direct approaches to few-shot learning.

Like many recent frameworks for multitask- and meta-learning, our approach
consists of three phases: a pretraining phase, a concept-learning phase, and an
evaluation phase. Here, the product of pretraining is a language interpretation
model that maps from descriptions to predictors (\eg image classifiers or
reinforcement learners).  Our thesis that language learning is a powerful,
general-purpose kind of pretraining, even for tasks that do not directly involve
language.

\begin{figure}[t!]
\centering
  \vspace{-.7em}
  \includegraphics[width=.9\columnwidth,clip,trim=0in 4in 3.3in 0in]{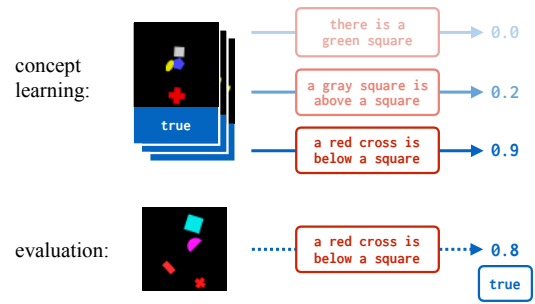}
  \vspace{-.2em}
  \caption{
    Example of our approach on a binary image classification task. We assume
    access to a pretrained language interpretation model that outputs the
    probability that an image matches a given description. To learn a new visual
    concept, we search in the space of natural language descriptions to maximize
    the interpretation model's score (top). The chosen description can 
    be used with the interpretation model to classify new images (bottom).
  }
  \label{fig:teaser}
  \vspace{-.5em}
\end{figure}

New concepts are learned by searching directly in the space of natural language
strings to minimize the loss incurred by the language interpretation model
(\autoref{fig:teaser}).  Especially on tasks that require the learner to
successfully model some high-level compositional structure shared by the
training examples, natural language hypotheses serve a threefold purpose: they
make it easier to discover these compositional concepts, harder to overfit to
few examples, and easier to understand inferred patterns.

Our approach can be implemented using a standard kit of neural components, and
is simple and general.  In a variety of settings, we find that the structure
imposed by a natural-language parameterization is helpful for efficient learning
and exploration.  The approach outperforms both multitask- and meta-learning
approaches that map directly from training examples to outputs by way of a
real-valued parameterization, as well as approaches that make use of natural
language annotations as an additional supervisory signal rather than an explicit
latent parameter.  The natural language concept descriptions inferred by our
approach often agree with human annotations when they are correct, and provide
an interpretable debugging signal when incorrect.  In short, by equipping models
with the ability to ``think out loud'' when learning, they become both more
comprehensible and more accurate.

\section{Background}
\label{sec:approach}

\begin{figure}
  \includegraphics[width=\columnwidth,clip,trim=0in 4.4in 1.8in 0.3in]{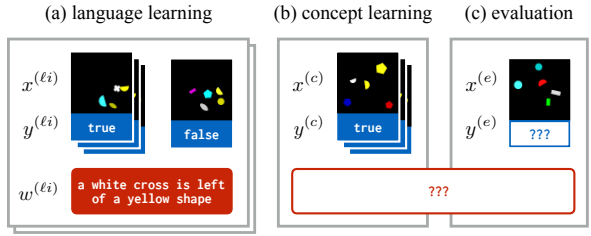}
  \caption{
    Formulation of the learning problem. Ultimately, we care about our model's
    ability to learn a concept from a small number of training examples (b) and
    successfully generalize it to held-out data (c). In this paper, concept
    learning is supported by a language learning phase (a) that makes use of
    natural language annotations on other learning problems. These annotations
    are not provided for the real target task in (b--c).
  }
  \label{fig:learning}
\end{figure}

Suppose we wish to solve an image classification problem like the one shown in
\autoref{fig:learning}b--c, mapping from images $\inp$ to binary labels $\outp$.
One straightforward way to do this is to solve a learning problem of the
following form:
\begin{equation}
  \label{eq:erm}
  \argmin_{\param\,\in\,\paramspace} ~~
  \sum_{\inp,\,\outp} \loss(\model(\inp;\,\param),\,\outp)
  ~,
\end{equation}
where $\loss$ is a loss function and $\model$ is a richly-parameterized class of
models (\eg convolutional networks) indexed by $\param$ (\eg weight matrices)
that map from images to labels.  Given a new image $\inp'$,
$\model(\inp';\,\param)$ can then be used to predict its label.

In the present work, we are particularly interested in \emph{few-shot} learning
problems where the number of $(\inp,\,\outp)$ pairs is small---on the order of
five or ten examples.  Under these conditions, directly solving \autoref{eq:erm}
is a risky proposition---any model class powerful enough to capture the true
relation between inputs and outputs is also likely to overfit. For few-shot
learning to be successful, extra structure must be supplied to the learning
algorithm.  Existing approaches obtain this structure by either carefully
structuring the hypothesis space or providing the learning algorithm with
additional kinds of training data. The approach we present in this paper
combines elements of both, so we begin with a review of existing work.

(Inductive) \textbf{program synthesis} approaches (\eg
\citealp{Gulwani11FlashFill}) reduce the effective size of the parameter space
$\paramspace$ by moving the optimization problem out of the continuous space of
weight vectors and into a discrete space of formal program descriptors (\eg
regular expressions or Prolog queries).  Domain-specific structure like version
space algebras \citep{Lau03VersionSpace} or type systems \citep{Kitzelmann06IFP}
can be brought to bear on the search problem, and the bias inherent in the
syntax of the formal language provides a strong prior.  But while program
synthesis techniques are powerful, they are also limited in their application: a
human designer must hand-engineer the computational primitives necessary to
compactly describe every learnable hypothesis. While reasonable for some
applications (like string editing), this is challenging or impossible in others
(like computer vision).

An alternative class of \textbf{multitask learning} approaches
\citep{Caruana98Multitask} attempt to import the relevant structure from other
learning problems rather than defining it manually (\autoref{fig:learning}a) .
Since we may not know \emph{a priori} what set of learning problems we
ultimately wish to evaluate on, it is useful to think of learning as taking
places in three phases:
\begin{enumerate}
  \item a \textbf{pretraining} (or ``meta-training'') phase that makes use of
    various different datasets $i$ with examples $\{ (\pinp{i}{1},
    \poutp{i}{1}), \ldots, (\pinp{i}{n}, \poutp{i}{n}) \}$
    (\autoref{fig:learning}a)
  \item a \textbf{concept-learning} phase in which the pretrained model is
    adapted to fit data $\{ (\cinp_1, \coutp_1), \ldots, (\cinp_n, \coutp_n) \}$
    for a specific new task (\autoref{fig:learning}b)
  \item an \textbf{evaluation} phase in which the learned concept is applied to
    a new input $\einp$ to predict $\eoutp$ (\autoref{fig:learning}c)
\end{enumerate}
In these approaches, learning operates over two collections of parameters:
shared parameters $\param$ and task-specific parameters $\tparam$. In
pretraining, multitask approaches find:
\begin{equation}
  \label{eq:multitask}
  \hspace{-1em}
  \argmin_{\param\,\in\,\reals^a,~\tparam^{(\ell i)}\,\in\,\reals^{b}} ~~
  \sum_{i,\,j}
  \loss\big(\model(\pinp{i}{j};\,\param,\,\tparam^{(\ell i)}),\,\poutp{i}{j}\big)
  ~.
\end{equation}
At concept learning time, they solve for:
\begin{equation}
  \label{eq:multitask-eval}
  \argmin_{~\tparam^{(c)}\,\in\,\reals^{b}} ~~
  \sum_j \loss\big(\model(\cinp_j;\,\param,\,\tparam^{(c)}),\,\coutp_j\big)
\end{equation}
on the new dataset, then make predictions for new inputs using $\model(\einp;
\,\param,\,\tparam^{(c)})$.

Closely related \textbf{meta-learning} approaches (\eg
\citealp{Schmidhuber87Thesis}; \citealp{Santoro16MANN};
\citealp{Vinyals16Matching}) make use of the same data, but collapse the inner
optimization over $\tparam^{(c)}$ and subsequent prediction of $\eoutp$ into a
single learned model.

\section{Learning with Language}

In this work, we are interested in developing a learning method that enjoys the
benefits of both approaches. In particular, we seek an intermediate language of
task representations that, like in program synthesis, is both expressive and
compact, but like in multitask approaches is learnable directly from training
data without domain engineering.  We propose to use natural language as this
intermediate representation.  We call our approach \linebreak \textbf{learning
with latent language} (\lc). 

Natural language shares many structural advantages with the formal languages
used in synthesis approaches: it is discrete, has a rich set of compositional
operators, and comes equipped with a natural description length prior. But it
also has a considerably more flexible semantics. And crucially, plentiful
annotated data exists for \emph{learning} this semantics: we cannot hand-write a
computer program to recognize a \emph{small dog}, but we can learn how to do it
from image captions. More basically, the set of primitive operators available in
language provides a strong prior about the kinds of abstractions that are useful
for natural learning problems.

Concretely, we replace the pretraining phase above with a
\textbf{language-learning} phase.  We assume that at language-learning time we
additionally have access to natural-language \textbf{descriptions} $\hint^{(\ell
i)}$ (\autoref{fig:learning}a, bottom).  We use these $\hint$ as
\emph{parameters}, in place of the task-specific parameters $\tparam$---that is,
we learn a language \textbf{interpretation} model
$\model(\inp;\,\param,\,\hint)$ that uses weights $\param$ to turn a description
$\hint$ into a function from inputs to outputs.  For the example in
\autoref{fig:learning}, $\model$ might be an image rating model
\citep{Socher14Multimodal} that outputs a scalar judgment $y$ of how well an
image $\inp$ matches a caption $\hint$.

Because these natural language parameters are observed at language-learning
time, we need only learn the real-valued shared parameters $\param$ used for
their interpretation (\eg the weights of a neural network that implements the
image rating model):
\begin{equation}
  \label{eq:multihint}
  \argmin_{\param\,\in\,\reals^a} ~~
  \sum_{i,\,j}
  \loss\big(\model(\pinp{i}{j};\,\param,\,\hint^{(\ell i)}),\,\poutp{i}{j}\big)
  ~.
\end{equation}
At concept-learning time, conversely, we solve only the part of the optimization
problem over natural language strings:
\begin{equation}
  \label{eq:multihint-eval}
  \argmin_{\hint'\,\in\,\Sigma^*} ~~
  \sum_j \loss\big(\model(\cinp_j;\,\param,\,\hint^{(c)}),\,\coutp_j\big)
  ~.
\end{equation}

This last step presents something of a challenge. When solving the corresponding
optimization problem, synthesis techniques can exploit the algebraic structure
of the formal language, while end-to-end learning approaches take advantage of
differentiability. Here we can't do either---the language of strings is
discrete, and whatever structure the interpretation function has is wrapped up
inside the black box of $f$.  Inspired by related techniques aimed at making
synthesis more efficient \citep{Devlin17RobustFill}, we use learning to help us
develop an effective optimization procedure for natural language parameters.

In particular, we simply use the language-learning datasets, consisting of pairs
$(\pinp{i}{j}, \poutp{i}{j})$ and descriptions $\hint_i$, to fit a reverse
\textbf{proposal} model, estimating:
\begin{equation}
\hspace{-1em}
\textstyle
  \argmax_\hparam \log q(\hint_i | \pinp{i}{1}, \poutp{i}{1}, \ldots, \pinp{i}{n}, \poutp{i}{n}; \hparam)
\end{equation}
where $q$ provides a (suitably normalized) approximation to the distribution of
descriptions given task data. In the running example, this proposal distribution
is essentially an image captioning model \citep{Donahue15LRCN}.  By sampling
from $q$, we expect to obtain candidate descriptions that are likely to obtain
small loss.  But our ultimate inference criterion is still the true model
$\model$: at evaluation time we perform the minimization in
\autoref{eq:multihint-eval} by drawing a fixed number of samples, selecting the
hypothesis $w^{(c)}$ that obtains the lowest loss, and using $\model(\einp;\,
\eta,\, w^{(c)})$ to make predictions.

What we have described so far is a generic procedure for equipping collections
of related learning problems with a natural language hypothesis space. In
Sections \ref{sec:shapes} and \ref{sec:re}, we describe how this procedure can
be turned into a concrete algorithm for supervised classification and sequence
prediction. In \autoref{sec:rl}, we describe how to extend these techniques to
reinforcement learning.

\section{Few-shot Classification}
\label{sec:shapes}

\begin{figure}[b]
  \centering
  \includegraphics[width=\columnwidth,clip,trim=0in 6.2in 5.35in 0in]{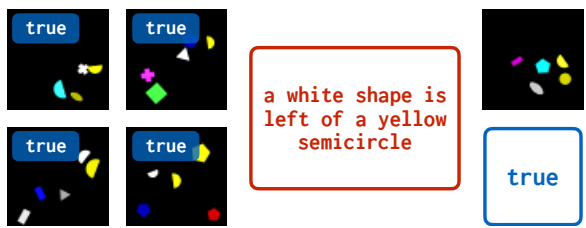}
  \caption{
    The few-shot image classification task. Learners are shown four positive
    examples of a visual concept (left) and must determine whether a fifth image
    matches the pattern (right). Natural language annotations are provided
    during language learning but must be inferred for concept learning.
  }
  \label{fig:shapes}
\end{figure}

We begin by investigating whether natural language can be used to support
high-dimensional few-shot classification.  Our focus is on visual reasoning
tasks like the one shown in \autoref{fig:shapes}. In these problems, the learner
is presented with four images, all positive examples of some visual concept like
\emph{a blue shape near a yellow triangle}, and must decide whether a fifth,
held-out image matches the same concept.

These kinds of visual reasoning problems have been well-studied in visual
question answering settings \citep{Johnson17CLEVR,Suhr17NLVR}. Our version of the
problem, where the input and output feature no text data, but a natural language
explanation must be inferred, is similar in spirit to the battery of visual
reasoning problems proposed by \citet{Raven36Matrices} and
\citet{Bongard68Problems}.

To apply the recipe in \autoref{sec:approach}, we need to specify an
implementation of the interpretation model $\model$ and the proposal model $q$.
We begin by computing representations of input images $\inp$. We start with a
pre-trained 16-layer VGGNet \citep{Simonyan14VGG}.  Because spatial information
is important for these tasks, we extract a feature representation from the final
convolutional layer of the network. This initial featurization is passed through
two fully-connected layers to form a final image representation, as follows: \\
\includegraphics[width=\columnwidth,clip,trim=0.22in 7in 4in 0in]{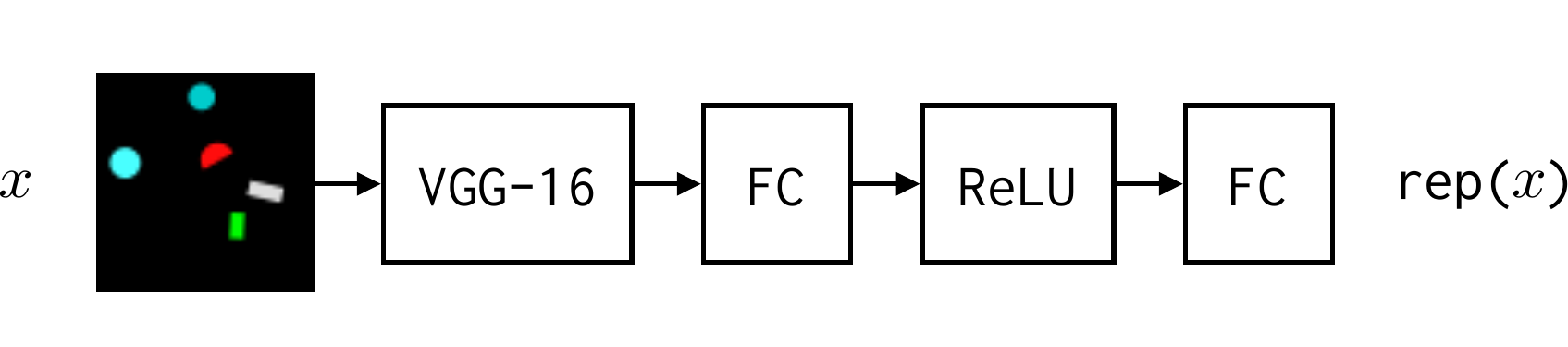} \\
Then we define interpretation and proposal models:\footnote{Suppressing 
global parameters $\param$ and $\lambda$ for clarity.}
\begin{align*}
  \model(\inp; \hint) &= \sigma\big(\cencode(\hint)^\top
  \crep(\inp)\big) \\
  q(\hint \mid \{\inp_j\}) &= \cdecode\big(\hint \mid
  {\textstyle\frac{1}{n}\sum}_j \crep(x_j) \big)
\end{align*}
The interpretation model $\model$ outputs the probability that $\inp$ is
assigned a positive class label, and is trained to maximize log-likelihood.
Because only positive examples are provided in each language learning set, the
proposal model $q$ can be defined in terms of inputs alone.  Details regarding
training hyperparameters, RNN implementations, etc.\ may be found in
\autoref{app:impl}. 

Our evaluation aims to answer two questions. First, does the addition of
language to the learning process provide any benefit over ordinary multitask or
meta-learning? Second, is it specifically better to use language as a hypothesis
space for concept learning rather than just an additional signal for
pretraining?  We use several baselines to answer these questions:
\begin{enumerate}
  \item \emph{Multitask}: a multitask baseline in which the definition of
    $\model$ above is replaced by $\sigma(\tparam_i^\top \texttt{rep}(x))$ for
    task-specific parameters $\tparam_i$ that are optimized during both
    pretraining and concept-learning.
  \item \emph{Meta}: a meta-learning baseline in which $\model$ is defined by
    $\sigma([\frac{1}{n}\sum_j \texttt{rep}(x_j)]^\top
    \texttt{rep}(x))$.\footnote{
      Many state-of-the-art approaches to meta-learning for classification (\eg
      \citealp{Snell17Proto}) are not well-defined for possibly-overlapping
      evaluation classes with only positive examples provideded. Here we have
      attempted to provide a robust implementation that is as close as possible
      to the other systems under evaluation. 
    }
  \item \emph{Meta+Joint}: as in \emph{Meta}, but the pretraining objective
    includes an additional term for predicting $q$ (discarded at
    concept-learning time).
\end{enumerate}

We report results on a dataset derived from the ShapeWorld corpus of
\citet{Kuhnle17ShapeWorld}. In this dataset the held-out image matches the
target concept 50\% of the time. In the validation and test folds, half of
learning problems feature a concept that also appears in the language learning
set (but with different exemplar images), while the other half feature both new
images and a new concept.  Images feature two or three distractor shapes
unrelated to the objects that define the target concept.  Captions in this
dataset were generated from DMRS representations using an HPS grammar
\citep{Copestake16DMRS}.  (Our remaining experiments use human annotators.) The
dataset contains a total of 9000 pretraining tasks and 1000 of each validation
and test tasks. More dataset statistics are provided in \autoref{app:data}.

\begin{table}[b!]
  \centering
  \footnotesize
  \begin{tabular}{lc>{\color{gray}}c>{\color{gray}}cccc}
    \toprule
    Model && Val (old) & Val (new) & Val && Test \\
    \cmidrule[0.5pt]{1-1} \cmidrule[0.5pt]{3-5} \cmidrule[0.5pt]{7-7}
    Random && 50 & 50 & 50 && 50 \\
    Multitask && 64 & 49 & 57 && 59 \\
    Meta && 63 & 62 & 62 && 64 \\
    Meta+Joint && 63 & 69 & 66 && 64 \\
    \lc (ours) && \bf 70 & \bf 72 & \bf 71 && \bf 70 \\[0.2em]
    \hdashline \\[-0.8em]
    \lc (oracle) && 77 & 80 & 79 && 78 \\
    \bottomrule
  \end{tabular}
  \caption{
    Evaluation on image classification. \emph{Val (old)} and \emph{Val (new)}
    denote subsets of the validation set that contain only previously-used and
    novel visual concepts respectively. \lc consistently outperforms alternative
    learning methods based on multitask learning, meta-learning, and
    meta-learning jointly trained to predict descriptions (\emph{Meta+Joint}).
    The last row of the table shows results when the model is given a
    ground-truth concept description rather than having to infer it from
    examples.
  }
  \label{tab:shapes-results}
\end{table}

\begin{figure}[b!]
  \centering
  \includegraphics[width=\columnwidth,clip,trim=0in 0in 3.7in 0in]{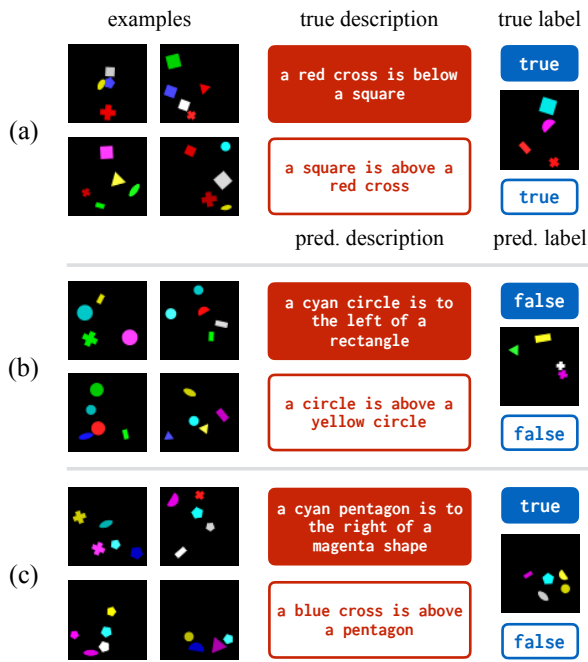}
  \\[-1em]
  \caption{
    Example predictions for image classification. The model achieves high accuracy
    even though predicted descriptions rarely match the ground
    truth. High-level structure like the presence of certain shapes or spatial
    relations is consistently recovered. Best viewed in color.
  }
  \label{fig:shapes-examples}
\end{figure}

Results are shown in \autoref{tab:shapes-results}. It can be seen that \lc
provides consistent improvements over the baselines, and that these improvements
are present both when identifying new instances of previously-learned concepts
and when discovering new ones.  Some example model predictions are shown in
\autoref{fig:shapes-examples}.  The model often succeeds in making correct
predictions, even though its inferred descriptions rarely match the ground
truth. Sometimes this is because of inherent ambiguity in the description
language (\autoref{fig:shapes-examples}a), and sometimes because the model is
able to rule out candidates on the basis of partial captions alone
(\autoref{fig:shapes-examples}b, where it is sufficient to recognize that the
target concept involves a \emph{circle}). More examples are provided in
\autoref{app:shape-examples}.

\section{Programming by Demonstration}
\label{sec:re}

\begin{figure}[b]
  \includegraphics[width=\columnwidth,clip,trim=0in 7in 5.05in 0.15in]{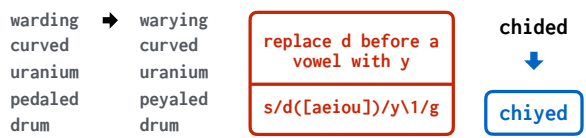}
  \caption{
    Example string editing task. Learners are presented with five examples
    of strings transformed according to some rule (left), and must apply an appropriate
    transformation to a sixth string (right). Language-learning annotations (center) may
    take the form of either natural language descriptions or regular
    expressions.
  }
  \label{fig:regex}
\end{figure}

Next we explore whether the same technique can be applied to tasks that involve
more than binary similarity judgments.  We focus on structured prediction:
specifically a family of string processing tasks.  In these tasks, the model is
presented with five strings being transformed according to some rule; it must
then apply an appropriate transformation to a sixth (\autoref{fig:regex}).
Learning proceeds more or less as in the previous section, with the following
definitions:
\begin{align*}
  &\crep(\inp, \outp) =
  \cencode([\inp, \outp]) \\
  &\model(\outp \mid \inp; \hint) = \\ 
  &\quad \cdecode\big(\outp \mid [\cencode(\inp), \cencode(\hint)] \big) \\
  &q(\hint \mid \{(x_j,\,y_j)\}) = \\ 
  &\quad \cdecode\big(\hint \mid {\textstyle\frac{1}{n}\sum}_j \crep(x_j, y_j) \big)
\end{align*}
Baselines are analogous to those for classification. 

While string editing tasks of the kind shown in \autoref{fig:regex} are popular
in both the programming by demonstration literature
\citep{Singh12StringTransform} and the semantic parsing literature
\citep{Kushman13Regex}, we are unaware of any datasets that support both
learning paradigms at the same time.  We have thus created a new dataset of
string editing tasks by (1) sampling random regular transducers, (2) applying
these transducers to collections of dictionary words, and (3) showing the
collected examples to Mechanical Turk users and asking them to provide a natural
language explanation with their best guess about the underlying rule.  The
dataset thus features both multi-example learning problems, as well as
structured and unstructured annotations for each target concept.  There are 3000
tasks for language learning and 500 tasks for each of validation and testing.
The human-generated data exhibits comparatively diverse explanatory strategies
and word choices; details are in \autoref{app:data}. Annotations are included in
the code release for this paper.

Results are shown in \autoref{tab:regex-results}. In these experiments, all
models that use descriptions have been trained on the natural language supplied
by human annotators.  While we did find that the Meta+Joint model converges
considerably faster than all the others, its final performance is somewhat lower
than the baseline Meta model.  As before, \lc outperforms alternative approaches
for learning directly from examples with or without descriptions.

\begin{table}[t]
  \vspace{-.2em}
  \centering
  \footnotesize
  \begin{tabular}{lcccc}
    \toprule
    Model && Val && Test \\
    \cmidrule{1-1} \cmidrule{3-3} \cmidrule{5-5}
    Identity && 18 && 18 \\
    Multitask && 54 && 50 \\
    Meta && 66 && 62 \\
    Meta+Joint && 63 && 59 \\
    \lc && \bf 80 && \bf 76 \\
    \bottomrule
  \end{tabular}
  \caption{
    Results for string editing. The reported number is the percentage of cases
    in which the predicted string exactly matches the reference. \lc is the best
    performing system; using language data for joint training rather than as a
    hypothesis space provides little benefit.
  }
  \label{tab:regex-results}
  \vspace{-.3em}
\end{table}

\begin{table}[b]
  \vspace{-.2em}
  \centering
  \footnotesize
  \begin{tabular}{lcccccc}
    \toprule
    \multirow{2}{*}{Annotations} && \multicolumn{2}{c}{Samples} &&
    \multicolumn{2}{c}{Oracle} \\
    && 1 & 100 && Ann. & Eval. \\
    \cmidrule{1-1} \cmidrule{3-4} \cmidrule{6-7}
    None (Meta) && \it 66 & -- && -- & -- \\
    Natural language && 66 & \it 80 && 75 & -- \\
    Regular expressions && 60 & 76 && 88 & 90 \\
    \bottomrule
  \end{tabular}
  \caption{
    Inference and representation experiments for string editing. Italicized
    numbers correspond to entries in \autoref{tab:regex-results}.  Allowing the
    model to use multiple samples rather than the 1-best decoder output
    substantially improves performance. The full model does better with inferred
    natural language descriptions than either regular expressions or
    ground-truth natural language.
  }
  \label{tab:regex-oracle}
\end{table}

\begin{figure}
  \includegraphics[width=\columnwidth,clip,trim=0.1in 0.3in 1.5in 0in]{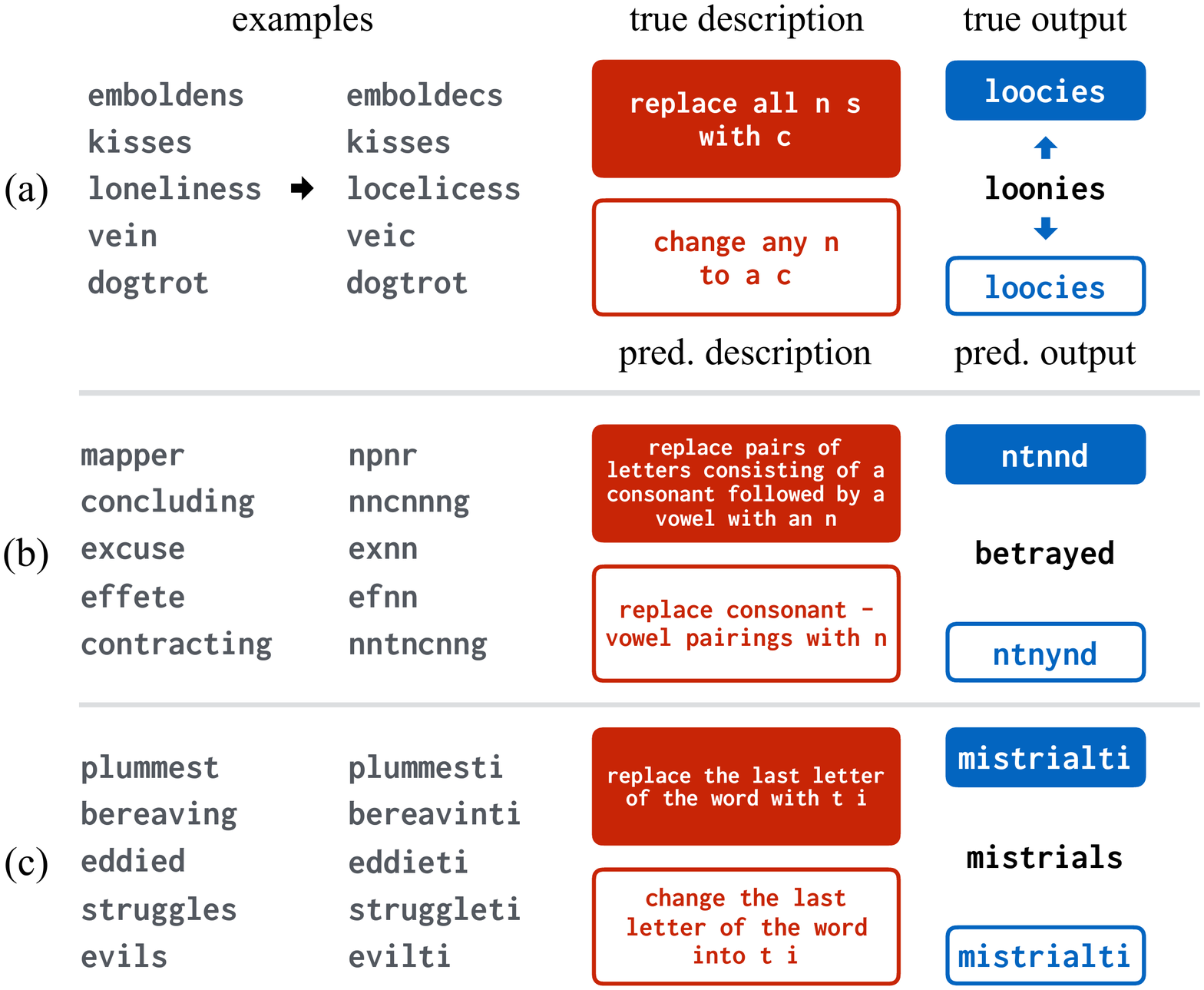}
  \caption{Example predictions for string editing.}
  \label{fig:regex-examples}
  \vspace{-.8em}
\end{figure}

Because all of the transduction rules in this dataset were generated from known
formal descriptors, these tasks provide an opportunity to perform additional
analysis comparing natural language to more structured forms of annotation
(since we have access to ground-truth regular expressions) and more conventional
synthesis-based methods (since we have access to a ground-truth regular
expression execution engine).  We additionally investigate the effect of the
number of samples drawn from the proposal model. These results are shown in
\autoref{tab:regex-oracle}.

A few interesting facts stand out. Under the ordinary evaluation condition (with
no ground-truth annotations provided), language-learning with natural language
data is actually better than language-learning with regular expressions. This
might be because the extra diversity helps the model figure out the relevant
axes of variation and avoid overfitting to individual strings. Allowing the
model to do its own inference is also better than providing ground-truth natural
language descriptions, suggesting that it is actually better at generalizing
from the relevant concepts than our human annotators (who occasionally write
things like \emph{I have no idea} for the inferred rule). Unsurprisingly, with
ground truth REs (which unlike the human data are always correct) we can do
better than any of the models that have to do inference. Coupling our inference
procedure with an oracle RE evaluator, we essentially recover the
synthesis-based approach of \citet{Devlin17RobustFill}. Our findings are
consistent with theirs: when a complete and accurate execution engine is
available, there is no reason not to use it. But we can get almost 90\% of the
way there with an execution model learned from scratch. Some examples of model
behavior are shown in \autoref{fig:regex-examples}; more may be found in
\autoref{app:regex-examples}.

\section{Policy Search}
\label{sec:rl}

\begin{figure}[b!]
  \centering
  \includegraphics[width=.8\columnwidth,clip,trim=.5in 15in 17.5in .5in]{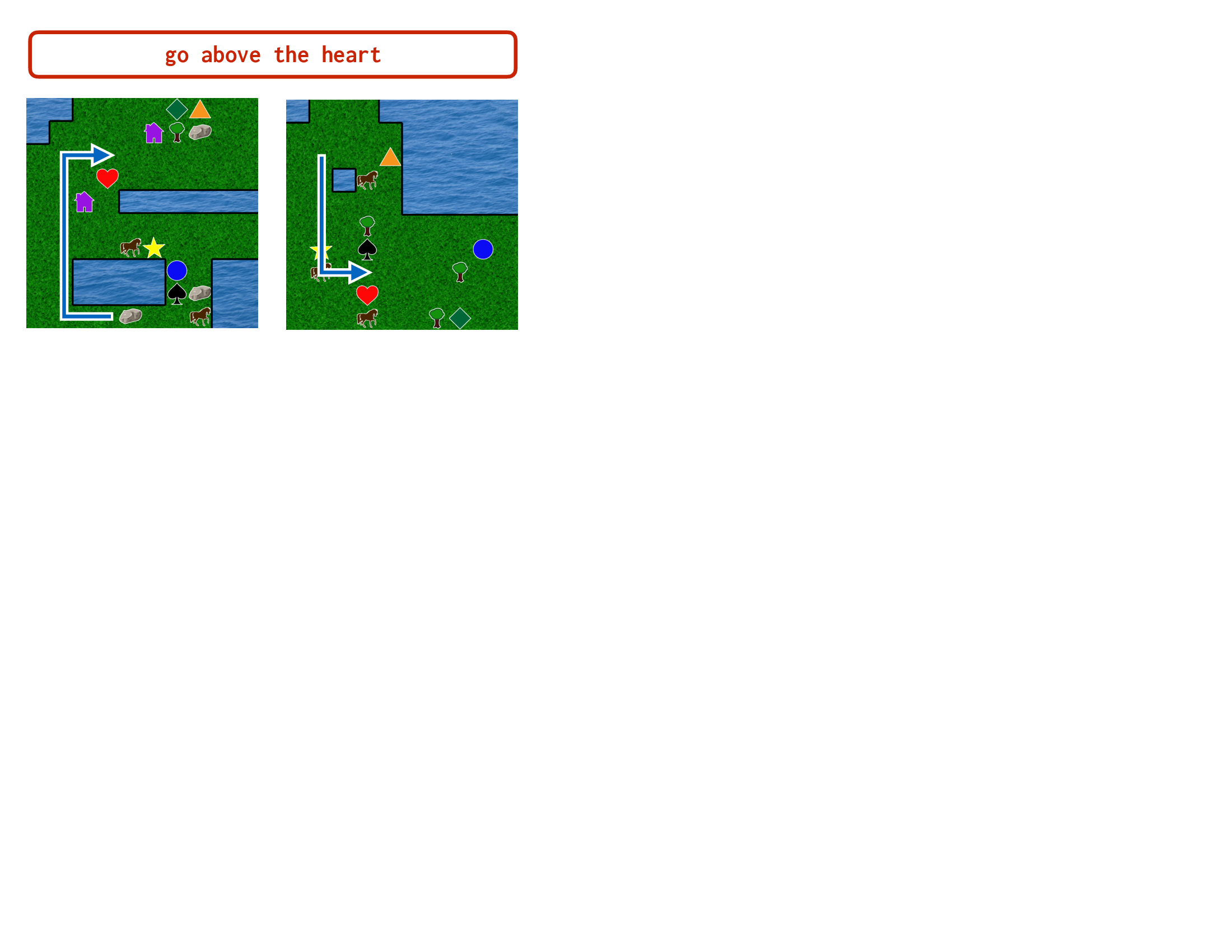}
  \caption{
    Example treasure hunting task: the agent is placed in a random environment
    and must collect a reward that has been hidden at a consistent offset with
    respect to some landmark. At language-learning time, natural language
    instructions and expert policies are additionally provided. The agent must
    both learn primitive navigation skills, like avoiding water, as well as the
    high-level structure of the reward functions for this domain.
  }
  \label{fig:nav}
\end{figure}

The previous two sections examined supervised settings where the learning signal
comes from few examples but is readily accessible. In this section, we move to a
set of reinforcement learning problems, where the learning signal is instead
sparse and time-consuming to obtain.  We evaluate on a collection of 2-D
treasure hunting tasks. These tasks require the agent to discover a rule that
determines the location of buried treasure in a large collection of environments
of the kind shown in \autoref{fig:nav}.  To recover the treasure, the agent must
navigate (while avoiding water) to its goal location, then perform a
\texttt{DIG} action. At this point the episode ends; if the treasure is located
in the agent's current position, it receives a reward, otherwise it does not.
In every task, the treasure has consistently been buried at a fixed position
relative to some landmark (like the heart in \autoref{fig:nav}). Both the offset
and the identity of the target landmark are unknown to the agent, and the
location landmark itself varies across maps.  Indeed, there is nothing about the
agent's observations or action space to suggest that landmarks and offsets are
even the relevant axis of variation across tasks, but this structure is made
clear in the natural language annotations. The high-level structure of these
tasks is similar to one used by \citet{Hermer01RelationalLanguageChild} to study
concept learning in humans.

The interaction between language and learning in these tasks is rather different
than in the supervised settings. In the supervised case, language served mostly
as a guard against overfitting, and could be generated conditioned on a set of
pre-provided concept-learning observations. Here, agents are free to interact
with the environment as much as they need, but receive observations only during
interaction. Thus our goal here will be to build agents that can \emph{adapt
quickly} to new environments, rather than requiring them to immediately perform
well on held-out data.

Why should we expect \lc to help in this setting? In reinforcement learning, we
typically encourage our models to explore by injecting randomness into either
the agent's action space or its underlying parameterization. But most random
random policies exhibit nonsensical behaviors; as a result, it is quite
inefficient both to sample in the space of network weights and to perform policy
optimization from a random starting point. Our hope is that when parameters are
instead chosen from within a structured family, a stochastic search in this
structured space will only ever consider behaviors corresponding to some
reasonable final policy, and in this way discover good behavior much faster than
ordinary RL.

Here the interpretation model $\model$ describes a policy that chooses actions
conditioned on the current environment state and its linguistic
parameterization. As the agent initially has no observations at all, we simply
design the proposal model to generate unconditional samples from a prior over
descriptions. 
Taking $\inp$ to be an agent's current observation of the environment state, we
define a state representation network:\\
\includegraphics[width=\columnwidth,clip,trim=0.22in 7in 4in 0in]{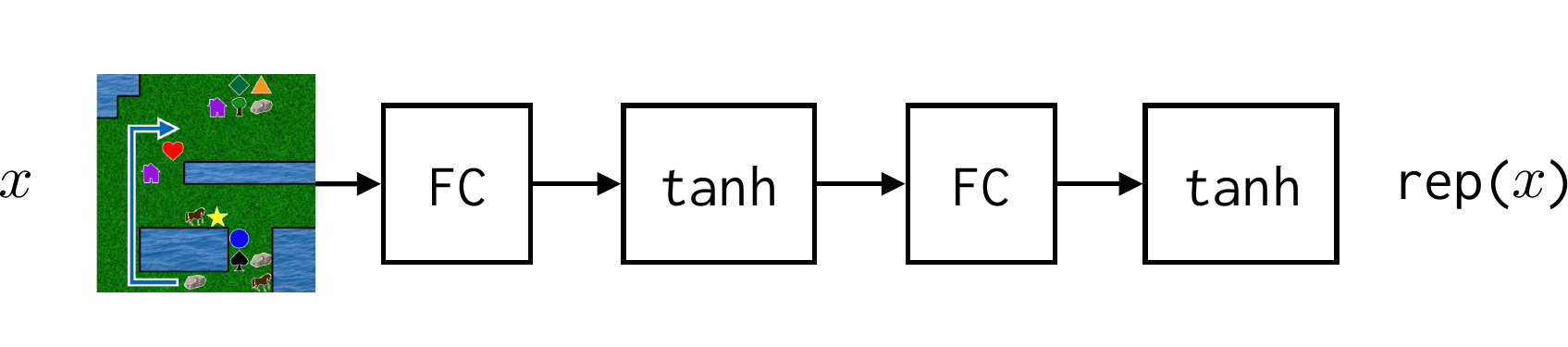} \\
and models:
\begin{align*}
  \model(a \mid \inp;\,\hint) &\propto \cencode(\hint)^\top\ W_a\ \crep(\inp) \\
  q(\hint) &= \cdecode(\hint)
\end{align*}
This parameterization assumes a discrete action space, and assigns to each
action a probability proportional to a bilinear function of the encoded
description and world state. $\model$ is effectively an instruction following
model of a kind well-studied in the natural language processing literature
\citep{Branavan09PG}; the proposal model allows it to generate its own
instructions without external direction.

To learn, we sample a fixed number of descriptions $\hint$ from $q$. For each
description, we sample multiple rollouts of the policy it induces to obtain an
estimate of the average reward it obtains. Finally, we can take the
highest-scoring description and perform additional fine-tuning of its induced
policy.

\begin{figure}
  \includegraphics[width=\columnwidth,clip,trim=.2in .3in 0in 0in]{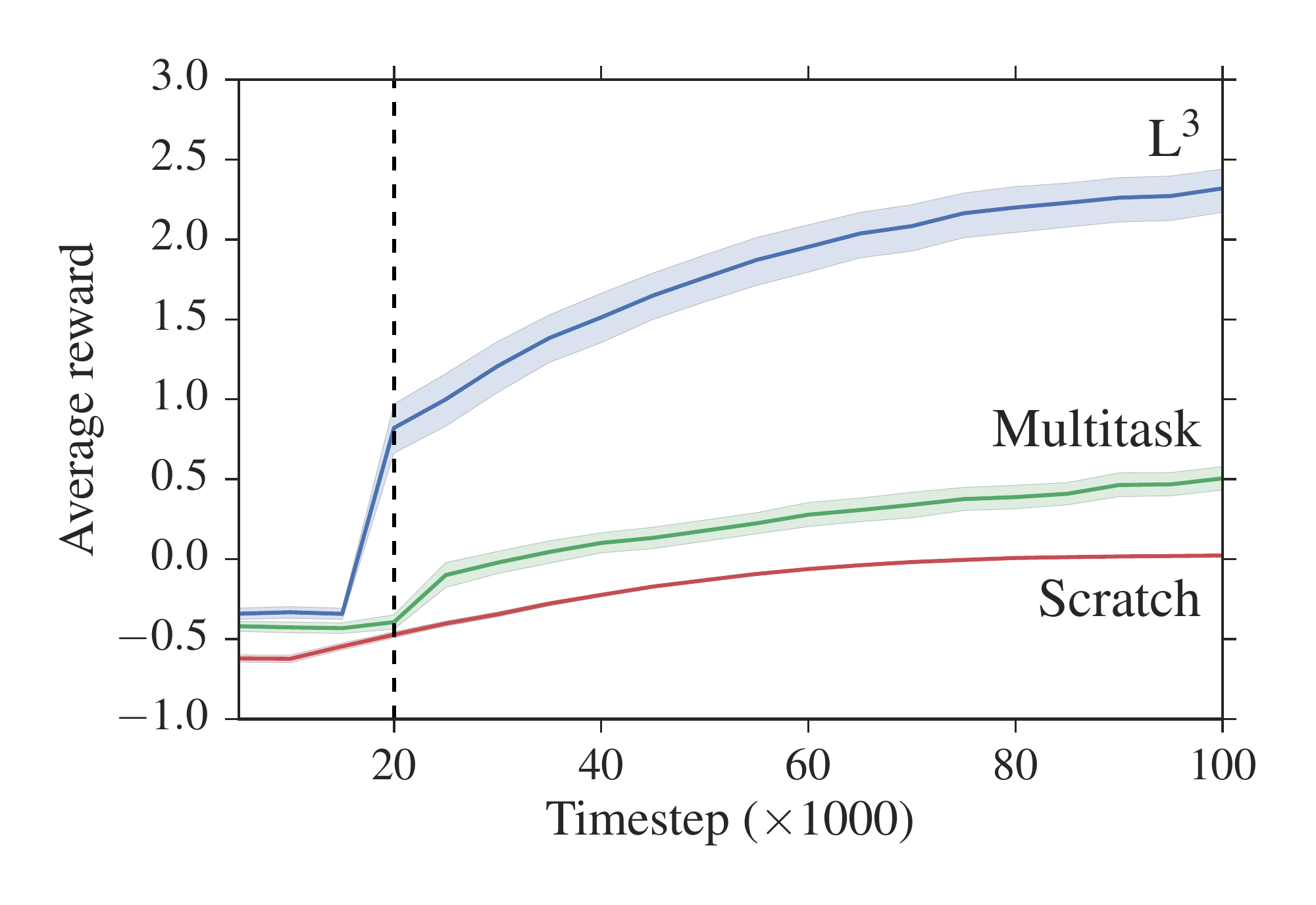}
  \caption{
    Learning curves for treasure hunting. These show the average reward obtained
    by each learning algorithm across multiple evaluation environments, after
    language learning has already taken place.  \emph{Multitask} learns a
    separate embedding for each task, while \emph{Scratch} trains on every task
    individually. \lc rapidly discovers high-scoring policies in most
    environments. The dashed line indicates the end of the concept-learning
    phase; subsequent performance comes from fine-tuning. The max possible
    reward for this task is 3 points.  Error bands shows 95\% confidence
    intervals for mean performance.
  }
  \label{fig:nav-results}
\end{figure}

At language-learning time, we assume access to both natural language
descriptions of these target locations provided by human annotators, as well as
expert policies for navigating to the location of the treasure. The multitask
model we compare to replaces these descriptions with trainable task
embeddings.\footnote{In the case of RL in particular, the contribution from \lc
are orthogonal to those of meta-learning---one could imagine using a technique
like RL$^2$ \citep{Duan16RL2} to generate candidate descriptions more
efficiently, or use \textsc{maml} \citep{Finn17MAML} rather than zero-shot
reward as the training criterion for the interpretation model.} The learner is
trained from task-specific expert policies using DAgger \citep{Ross11DAgger}
during the language-learning phase, and adapts to individual environments using
``vanilla'' policy gradient \citep{Williams92Reinforce} during the concept
learning phase.

The environment implementation and linguistic annotations are in this case
adapted from a natural language navigation dataset originally introduced by
\citet{Janner17Maps}. In our version of the problem (\autoref{fig:nav}), the
agent begins each episode in a random position on a randomly-chosen map and must
attempt to obtain the treasure. Some relational concepts describing target
locations are reused between language learning and concept-learning phases, but
the environments themselves are distinct. For language learning the agent has
access to 250 tasks, and is evaluated on an additional set of 50. 

Averaged learning curves for held-out tasks are shown in
\autoref{fig:nav-results}.  As expected, reward for the \lc model remains low
during the initial exploration period, but once a description is chosen the
score improves rapidly. Immediately \lc achieves better reward than the
multitask baseline, though it is not perfect; this suggests that the
interpretation model is somewhat overfit to the pretraining environments.
However, after additional fine-tuning even better results are rapidly obtained.
Example rollouts are visualized in \autoref{app:nav-examples}.  These results
show that the model has used the structure provided by language to \emph{learn}
a better representation space for policies---one that allows it to sample from a
distribution over interesting and meaningful behaviors rather than sequences of
random actions.

\section{Other Related Work}

This is the first approach we are aware of to frame a general learning problem
as optimization over a space of natural language strings.  However, many
closely-related ideas have been explored in the literature. String-valued latent
variables are widely used in language processing tasks ranging from
morphological analysis \citep{Dreyer09StringPGM} to sentence compression
\citep{Miao16LatentLanguage}.  Natural language annotations have been used in
conjunction with training examples to guide the discovery of logical
descriptions of concepts \citep{Ling17RationaleGeneration,Srivastava17Concepts},
and used as an auxiliary loss for training \citep{Frome13Devise}, analogously to
the Meta+Joint baseline in this paper.  Structured language-like annotations
have been used to improve learning of generalizable structured policies
\citep{Andreas17Sketches,Denil17ProgrammableAgent}. Finally, natural language
instructions available at \emph{concept-learning} time (rather than
language-learning time) have been used to provide side information to
reinforcement learners about high-level strategy \citep{Branavan11Civ},
environment dynamics \citep{Narasimhan17RLTransfer} and exploration
\citep{Harrison17RLGuiding}.

\section{Conclusion}

We have presented an approach for optimizing models in a space parameterized by
natural language.  Using standard neural encoder--decoder components to build
models for representation and search in this space, we demonstrated that our
approach outperforms strong baselines on classification, structured prediction
and reinforcement learning tasks. We believe that these results suggest the
following general conclusions:

\emph{Language encourages compositional generalization}: standard deep learning
architectures are good at recognizing new instances of previously-encountered
concepts, but not always at generalizing to new ones. By forcing decisions to
pass through a linguistic bottleneck in which the underlying compositional
structure of concepts is explicitly expressed, stronger generalization becomes
possible. 
\linebreak \indent \emph{Language simplifies structured exploration}:
relatedly, linguistic scaffolding can provide dramatic advantages in problems
like reinforcement learning that require exploration: models with latent
linguistic parameterizations can sample in this space, and thus limit
exploration to a class of behaviors that are likely \emph{a priori} to be
goal-directed and interpretable.

And generally, \emph{language can help learning}. In multitask settings, it can
even improve learning on tasks for which no language data is available at
training or test time.  While some of these advantages are also provided by
techniques like program synthesis that are built on top of formal languages,
natural language is at once more expressive and easier to obtain than formal
supervision. We believe this work hints at broader opportunities for using
naturally-occurring language data to improve machine learning for tasks of all
kinds.

\clearpage

%% file: appendix.tex
\clearpage

\appendix

\section{Model and Training Details}
\label{app:impl}

In all models, RNN encoders and decoders use gated recurrent units
\citep{Cho14GRU}.

\paragraph{Few-shot classification}
Models are trained with the \textsc{adam} optimizer \citep{Kingma14Adam} with a
step size of 0.0001 and batch size of 100. The number of pretraining iterations
is tuned based on subsequent concept-learning performance on the development
set.  Neural network hidden states, task parameters, and word vectors are all
of size 512. 10 hypotheses are sampled during for each evaluation task in the
concept-learning phase.

\paragraph{Programming by demonstration}
Training as in the classification task, but with a step size of 0.001.  Hidden
states are of size 512, task parameters of size 128 and word vectors of size
32. 100 hypotheses are sampled for concept learning.

\paragraph{Policy search}
DAgger \citep{Ross11DAgger} is used for pre-training and vanilla policy gradient
\citep{Williams92Reinforce} for concept learning. Both learning algorithms use
\textsc{adam} with a step size of 0.001 and a batch size of 5000 samples.  For
imitation learning, rollouts are obtained from the expert policy on a schedule
with probability $0.95^t$ (for $t$ the current epoch).  For reinforcement
learning, a discount of 0.9 is used. Because this dataset contains no
development data, pretraining is run until performance on the pretraining tasks
reaches a plateau. Hidden states and task embeddings are of size 64.  100
hypotheses are sampled for concept learning, and 1000 episodes (divided evenly
among samples) are used to estimate hypothesis quality before fine-tuning.

\section{Dataset Information}
\label{app:data}

\paragraph{ShapeWorld}
This is the only fully-synthetic dataset used in our experiments.  Each scene
features 4 or 5 non-overlapping entities. Descriptions refer to spatial
relationships between pairs of entities identified by shape, color, or both.
There are 8 colors and 8 shapes. The total vocabulary size is only 30 words, but
the dataset contains 2643 distinct captions.  Descriptions are on average 12.0
words long.

\paragraph{Regular expressions}
Annotations were collected from Mechanical Turk users. Each user was presented
with the same task as the learner in this paper: they observed five strings
being transformed, and had to predict how to transform a sixth. Only after they
correctly generated the held-out word were they asked for a description of the
rule. Workers were additionally presented with hints like ``look at the
beginning of the word'' or ``look at the vowels''.  Descriptions are
automatically preprocessed to strip punctuation and ensure that every character
literal appears as a single token.

The regular expression data has a vocabulary of 1015 rules and a total of 1986
distinct descriptions. Descriptions are on average 12.3 words in length but as
long as 46 words in some cases.

\paragraph{Navigation}

The data used was obtained from \citet{Janner17Maps}. We created our own variant
of the dataset containing collections of related tasks. Beginning with the
``local'' tasks in the dataset, we generated alternative goal positions at fixed
offsets from landmarks as described in the main section of this paper.
Natural-language descriptions were selected for each task collection from the
human annotations provided with the dataset. The vocabulary size is 74 and the
number of distinct hints 446. The original action space for the environment is
also modified slightly: rather than simply reaching the goal cell (achieved with
reasonably high frequency by a policy that takes random moves), we require the
agent to commit to an individual goal cell and end the episode with a special
\texttt{DIG} action.

\paragraph{Data augmentation}

Due to their comparatively small size, a data augmentation scheme
\citep{Jia16Recombination} is employed for the regular expression and navigation
datasets. In particular, wherever a description contains a recognizable entity
name (\ie a character literal or a landmark name), a description template is
extracted. These templates are then randomly swapped in at training time on
other examples with the same high-level semantics. For example, the description
\emph{replace first b with e} is abstracted to \emph{replace first CHAR1 with
CHAR2}, and can subsequently be specialized to, e.g., \emph{replace first c with
d}. This templating is easy to implement because we have access to ground-truth
structured concept representations at training time.  If these were not
available it would be straightforward to employ an automatic template induction
system \citep{Kwiatkowski11TUBL} instead.

\onecolumn

\section{Examples: ShapeWorld}
\label{app:shape-examples}

\strut
(Examples in this and the following appendices were not cherry-picked.) \\
\strut

\footnotesize
\tt

\newcommand{\srow}[5]{
  \midrule
  \\[-.6em]
  \makecell[c]{
    \includegraphics[width=.50in]{fig/shapes/vis100/#1/ex_0.png}
    \includegraphics[width=.50in]{fig/shapes/vis100/#1/ex_1.png}\\
    \includegraphics[width=.50in]{fig/shapes/vis100/#1/ex_2.png}
    \includegraphics[width=.50in]{fig/shapes/vis100/#1/ex_3.png}
  } &
  \makecell[l]{#2 \\[2em] #3} &
  \makecell[c]{
  \includegraphics[width=.50in]{fig/shapes/vis100/#1/input.png}} &
  \makecell[c]{#4 \\[2em] #5} \\
  \\[-.7em]
}

\newcommand{\sbad}{\color{red}}

\noindent
\begin{tabular*}{\textwidth}{@{\extracolsep{\fill}}lllll}
  \midrule
  \\[-.7em]
  \makecell[c]{
    {\bf Positive examples:}\\[.5em]
    \includegraphics[width=.55in]{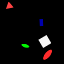}
    \includegraphics[width=.55in]{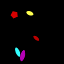}\\
    \includegraphics[width=.55in]{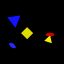}
    \includegraphics[width=.55in]{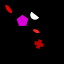}
  } &
  \makecell[l]{
    {\bf True description}:\\ a red ellipse is to the right of an ellipse \\[2em] 
    {\bf Inferred description:}\\ \sbad a red shape is to the right of a red semicircle
  } &
  \makecell[c]{
    {\bf Input:}\\[.5em]
  \includegraphics[width=.55in]{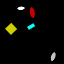}} &
  \makecell[c]{ {\bf True label:}\\ true \\[2em] {\bf Pred.\ label:}\\ true} \\
  \\[-.8em]

  \srow{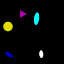}{
    a shape is below a white ellipse
  }{
    \sbad a white shape is to the left of a yellow ellipse
  }{false}{\sbad true}

  \srow{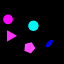}{
    a magenta triangle is to the left of a magenta pentagon
  }{
    a magenta triangle is to the left of a pentagon
  }{true}{true}

  \srow{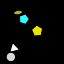}{
    a green pentagon is to the right of a yellow shape
  }{
    \sbad a green shape is to the right of a red semicircle
  }{false}{false}

  \srow{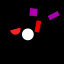}{
    a red circle is above a magenta semicircle
  }{
    \sbad a green triangle is above a red circle
  }{false}{\sbad true}

  \srow{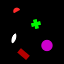}{
    a white ellipse is to the left of a green cross
  }{
    a green cross is to the right of a white ellipse
  }{true}{true}
  \midrule
\end{tabular*}
\newpage

\section{Examples: Regular Expressions}
\label{app:regex-examples}
\scriptsize

\newcommand{\rrow}[7]{
  \midrule
  \\[-.7em]
  \makecell[l]{#1} & 
  \makecell[l]{#2} & 
  \makecell[l]{#3 \\[2em] #4} & 
  \makecell[l]{#5} &
  \makecell[l]{#6 \\[2em] #7} \\
  \\[-.7em]
}
\newcommand{\rbad}{\color{red}}

\strut

\noindent
\begin{tabular*}{\textwidth}{@{\extracolsep{\fill}}lllll}
  \rrow{
    {\bf Example in:} \\
    mediaeval \\
    paneling \\
    wafer \\
    conventions \\
    handsprings
  }{
    {\bf Example out:} \\
    ilediaeval \\
    ilaneling \\
    ilafer \\
    ilonventions \\
    ilandsprings
  }{
    {\bf Human description:}\\
    leading consonant si replaced with i l
  }{
    {\bf Inferred description:}\\
    first consonant of a word is replaced with i l
  }{{\bf Input:}\\ chaser}{{\bf True out:}\\ ilhaser}{\bf{Pred.\ out:}\\ilhaser}

  \rrow{
    uptakes \\
    pouching \\
    embroidery \\
    rebelliousness \\
    stoplight
  }{
    uptakes \\
    punuching \\
    embrunidery \\
    rebelliunusness \\
    stunplight
  }{
    replace every o with u n
  }{
    change all o to u n
  }{regulation}{regulatiunn}{\rbad regulatinun}

  \rrow{
    fluffiest \\
    kidnappers \\
    matting \\
    griping \\
    disagreements
  }{
    fluffiest \\
    kidnappers \\
    eeatting \\
    griping \\
    disagreeeeents
  }{
    the leter m is replaced by ee
  }{
    change every m to ee
  }{chartering}{chartering}{chartering}

  \rrow{
    clandestine \\
    limning \\
    homes \\
    lifeblood \\
    inflates
  }{
    clandqtine \\
    limning \\
    homq \\
    lifqlood \\
    inflatq
  }{
    \rbad e
  }{
    where e appears , replace it \\ and the following letter 
    with q
  }{gratuity}{gratuity}{gratuity}

  \rrow{
    fruitlessly \\
    sandier \\
    washers \\
    revelries \\
    dewlaps
  }{
    fruitlessly \\
    sandier \\
    washemu \\
    revelrimu \\
    dewlamu
  }{
    if the word ends with an s , replace \\ the last two letters with m u
  }{
    \rbad change last to m u if consonant
  }{prompters}{promptemu}{promptemu}

  \rrow{
    ladylike \\
    flintlocks \\
    student \\
    surtaxes \\
    bedecks
  }{
    ladylike \\
    flintlocknl \\
    studennl \\
    surtaxenl \\
    bedecknl
  }{
    ending consonant is replaced with n l
  }{
    \rbad drop last two and add n l
  }{initials}{initialnl}{initialnl}

  \rrow{
    porringer \\
    puddling \\
    synagog \\
    curtseying \\
    monsieur
  }{
    porringeer \\
    puddlinge \\
    synageoge \\
    curtseyinge \\
    monsieur
  }{
    add e next to letter g
  }{
    \rbad when a letter is preceded by a g , \\ 
    \rbad e is added after that letter
  }{rag}{rage}{rage}

  \rrow{
    trivializes \\
    tried \\
    tearfully \\
    hospitalize \\
    patronizing
  }{
    trivializes \\
    tried \\
    gxarfully \\
    gxspitalize \\
    gxtronizing
  }{
    replace the 1st 2 letters of the word with a g x \\ if the word 
    begins with a consonant then a vowel
  }{
    if the second letter is a vowel , replace the \\ first two letters with g x
  }{landlords}{gxndlords}{gxndlords}

  \rrow{
    microseconds \\
    antiviral \\
    flintlock \\
    appreciable \\
    stricter
  }{
    microsecnyr \\
    antiviral \\
    flintloyr \\
    appreciabyr \\
    stricter \\
  }{
    \rbad replace consonants with y r
  }{
    \rbad the last two letters are replaced by y r
  }{exertion}{exertion}{\rbad exertiyr}
  
  \midrule
\end{tabular*}

\newpage
\section{Examples: Navigation}
\label{app:nav-examples}

\strut\\[-0.5em]
{\noindent\normalsize \rmfamily White breadcrumbs show the path taken by the
agent.}\\
\strut

\newcommand{\nrow}[3]{
  \midrule
  \\
  \makecell[l]{ #1 \\[1em] #2 } &
  \makecell[c]{\includegraphics[width=1in]{fig/nav/vis/#3/map_0}} &
  \makecell[c]{\includegraphics[width=1in]{fig/nav/vis/#3/map_1}} &
  \makecell[c]{\includegraphics[width=1in]{fig/nav/vis/#3/map_2}} \\
  \\
}

\noindent
\begin{tabular*}{\textwidth}{@{\extracolsep{\fill}}llll}
  \nrow{
    {\bf Human description:}\\ move to the star
  }{
    {\bf Inferred description:}\\ reach the star cell
  }{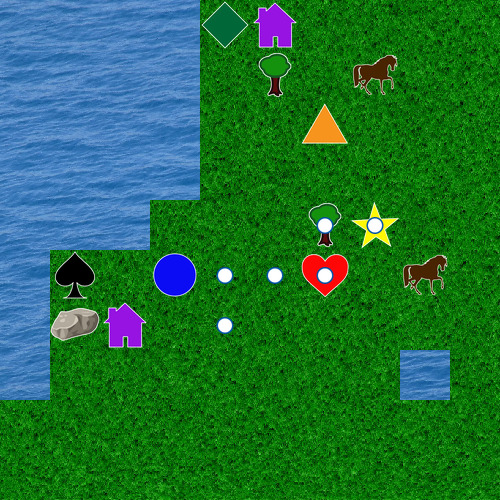}
  \nrow{reach square one right of triangle}{reach cell to the right of the triangle}{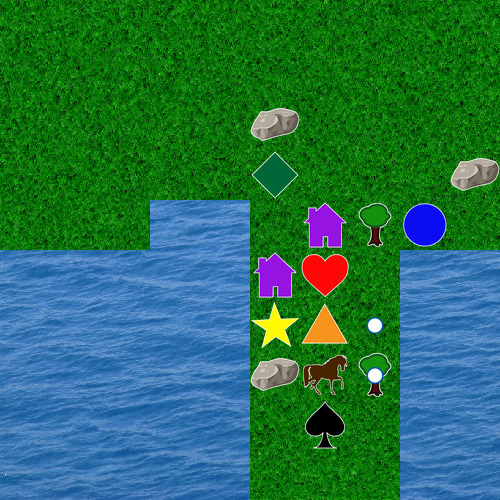}
  \nrow{reach cell on left of triangle}{reach square left of triangle}{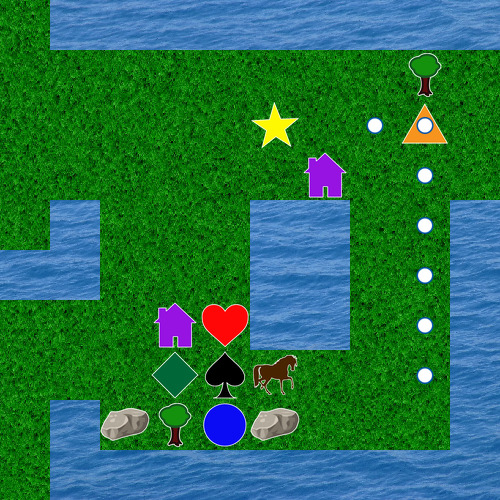}
  \nrow{reach spade}{go to the spade}{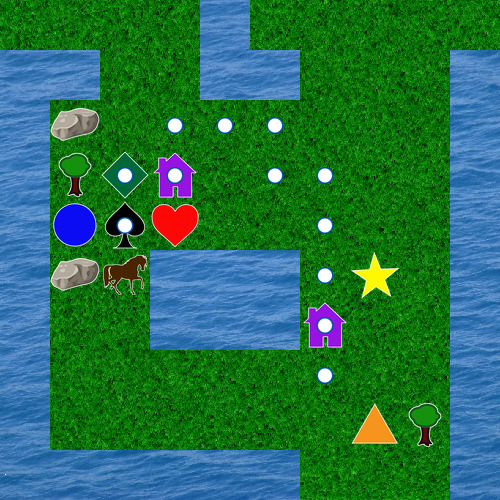}
  \nrow{left of the circle}{go to the cell to the left of the circle}{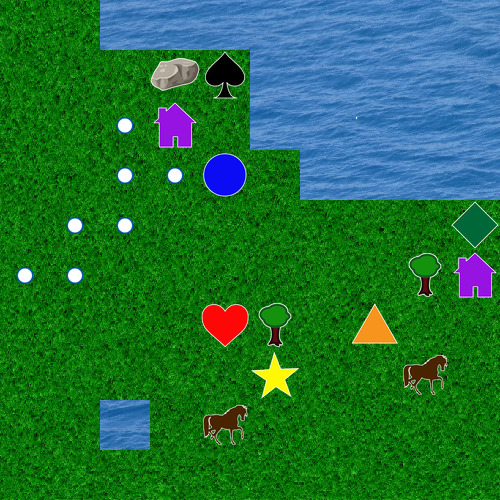}
  \nrow{reach cell below the circle}{reach cell below circle}{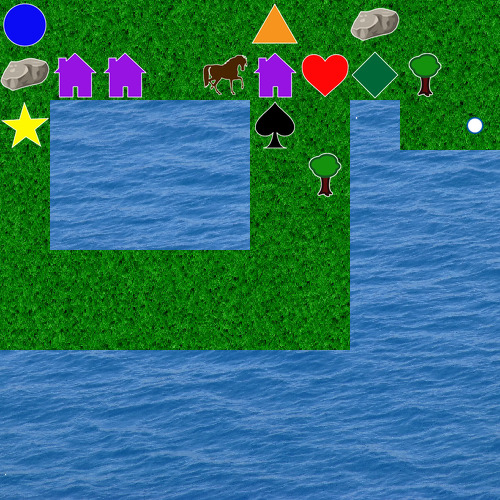}
  \midrule
\end{tabular*}